\documentclass{article}

\usepackage{ijcai25}

\usepackage{times}
\usepackage{soul}
\usepackage{url}
\usepackage[hidelinks]{hyperref}
\usepackage[utf8]{inputenc}
\usepackage[small]{caption}
\usepackage{graphicx}
\usepackage{amsmath}
\usepackage{amsthm}
\usepackage{amssymb}
\usepackage{booktabs}
\usepackage{algorithm}
\usepackage{algorithmic}
\usepackage[switch]{lineno}
\usepackage{multirow}
\usepackage{xcolor}
\usepackage{CJKutf8}
\definecolor{blueColor}{rgb}{0.5,0.5,1} 

\definecolor{orgColor}{rgb}{1,0.5,0}

\newcommand{\LG}{MM-LG}
\usepackage{xcolor}
\usepackage{colortbl}

\title{Extracting Multimodal Learngene in CLIP: Unveiling the Multimodal Generalizable Knowledge}

\author{
 Ruiming Chen, Junming Yang, Shiyu Xia, Xu Yang\thanks{
    Co-corresponding auther.}, Jing Wang, Xin Geng \footnotemark[1]
    \affiliations
  School of Computer Science and Engineering, Southeast University, Nanjing 210096, China\\
  Key Laboratory of New Generation Artificial Intelligence Technology and Its Interdisciplinary Applications (Southeast University), Ministry of Education, China
\emails
  \{220232251, jmingyang, shiyu\_xia, xuyang\_palm, wangjing91, xgeng\}@seu.edu.cn\\
}

\begin{document}

\maketitle

\begin{abstract}
    CLIP~(Contrastive Language-Image Pre-training) has attracted widespread attention for its multimodal generalizable knowledge, which is significant for downstream tasks. However, the computational overhead of a large number of parameters and large-scale pre-training poses challenges of pre-training a different scale of CLIP. \textit{Learngene} extracts the generalizable components termed as \textit{learngene} from an ancestry model and initializes diverse descendant models with it. Previous Learngene paradigms fail to handle the generalizable knowledge in multimodal scenarios. In this paper, we put forward the idea of utilizing a multimodal block to extract the multimodal generalizable knowledge, which inspires us to propose \textbf{MM-LG} (\textbf{M}ulti\textbf{m}odal \textbf{L}earn\textbf{g}ene), a novel framework designed to extract and leverage generalizable components from CLIP. Specifically, we first establish multimodal and unimodal blocks to extract the multimodal and unimodal generalizable knowledge in a weighted-sum manner. Subsequently, we employ these components to numerically initialize descendant models of varying scales and modalities.
    Extensive experiments demonstrate MM-LG's effectiveness, which achieves performance gains over existing learngene approaches~(\textit{e.g.},+3.1\% on Oxford-IIIT PET and +4.13\% on Flickr30k) and comparable or superior results to the pre-training and fine-tuning paradigm~(\textit{e.g.},+1.9\% on Oxford-IIIT PET and +3.65\% on Flickr30k). Notably, MM-LG requires only around 25\% of the parameter storage while reducing around 2.8$\times$ pre-training costs for diverse model scales compared to the pre-training and fine-tuning paradigm, making it particularly suitable for efficient deployment across diverse downstream tasks.

\end{abstract}

\section{Introduction}
As a representative paradigm in multimodal research, CLIP (Contrastive Language-Image Pre-training)~\cite{radford2021learning} leverages 400 million large-scale image-text pairs for contrastive pre-training. 
It achieves powerful representation capabilities and effectively integrates both modalities, demonstrating strong cross-modal generalization ability. 
These characteristics are manifested in excelling in various downstream tasks, including zero-shot image classification~\cite{pmlr-v202-novack23a}, cross-modal retrieval~\cite{10.1145/3626772.3657678} and image captioning~\cite{Mokady2021ClipCapCP}. 
Despite these advancements, an excessive number of parameters to be deployed has become a barrier for edge devices~\cite{Yang2024CLIPCIDEC,yang2024clip}. Besides, if a different model scale is required, the repetitive computational requirements of pre-training are extremely time-consuming and computationally expensive.
\begin{figure}[t]
    \centering
    \includegraphics[width=0.48\textwidth]{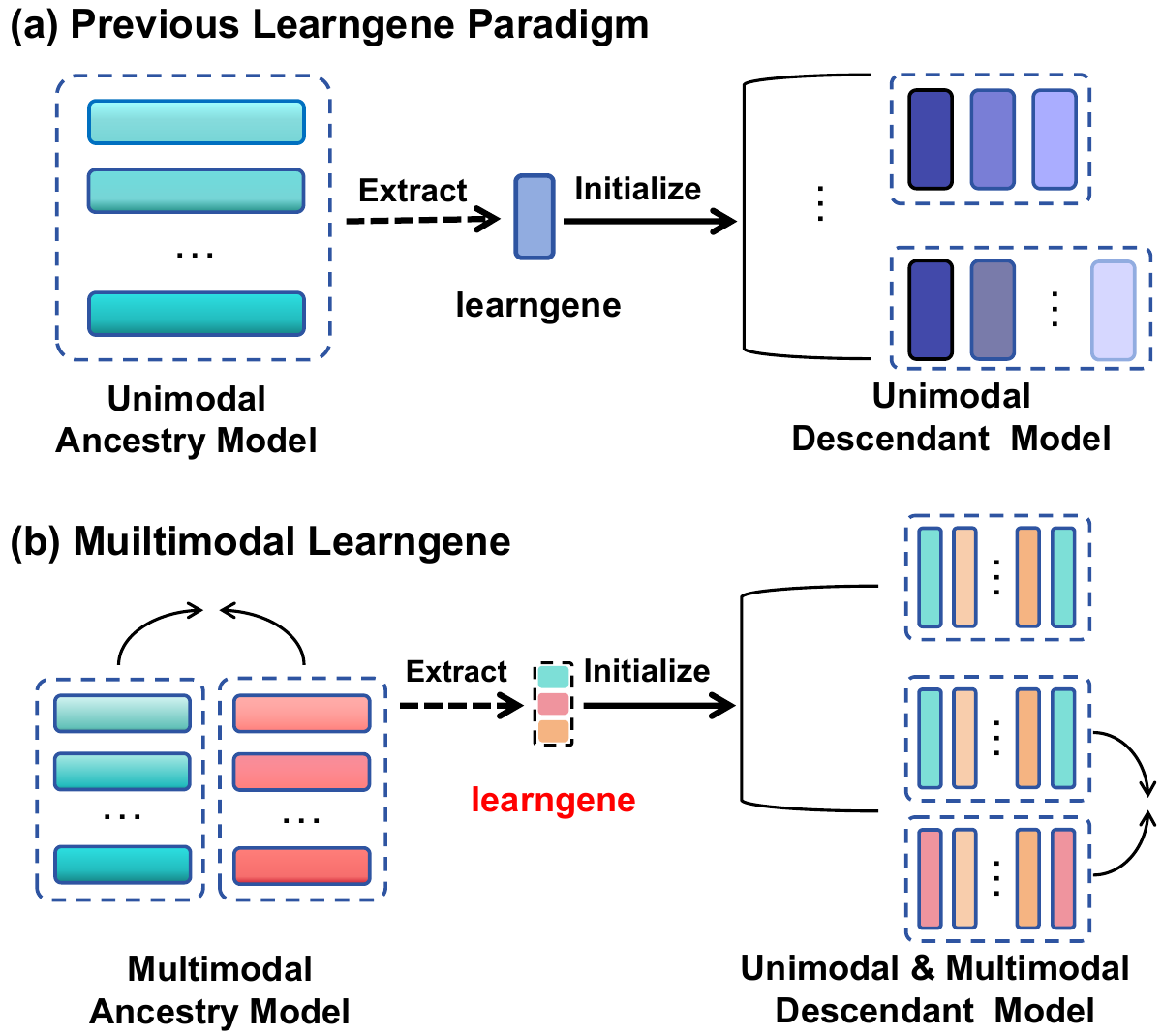}
    \caption{(a) Learngene paradigm comprises two stages. The first stage is to extract the compact learngene from an ancestry model. Secondly, diverse descendant models are initialized with it. (b) Multimodal Learngene first extracts learngene, which is generalizable across modalities, and subsequently initializes multimodal and unimodal descendant models of varying scales with it.}
    \label{fig:intro}
\end{figure}

\textit{Learngene}, first proposed by~\cite{wang2022learngene}, 
preserves the generalizable component of the pre-trained model and initializing diverse downstream models with it, as depicted in Fig.~\ref{fig:intro}(a).
The first stage is to extract the generalizable component, termed as \textbf{learngene}, from a large pre-trained model (referred to as the ancestry model).
In the second stage, this extracted learngene is utilized to initialize diverse models (referred to as the descendant model) for downstream tasks.
This approach effectively provides an effective solution for facilitating efficient knowledge transfer.


Recent studies have extensively investigated and validated the effectiveness of the Learngene paradigm~\cite{wang2023learngene,xia2024exploring,Xia2023Linear,feng2024wave}.
He-LG~\cite{wang2022learngene} proposed a gradient-based approach to extract learngene, which is used to initialize descendant models stacked with randomly initialized layers.
TLEG~\cite{Xia2023Linear} developed a linear extraction method specifically for Transformer-based ancestry models, employing numerical initialization techniques for descendant models.
However, for these existing works, there is a lack of research into leveraging multimodal generalizable knowledge for learngene extraction within multimodal models.
To advance the exploration of generalizable learngene in multimodal architectures, we conduct our investigation on CLIP. As illustrated in Fig.~\ref{fig:intro} (b), to preserve the multimodal generalizable knowledge in CLIP explicitly, we seek to construct a multimodal block to handle it. However, it is insufficient to solely rely on the multimodal block since there is unimodal generalizable knowledge in CLIP. Thus unimodal blocks are established to store these components, represented as vision and language blocks. Furthermore, to effectively extract generalizable knowledge into these blocks from CLIP, an auxiliary model~\cite{Xia2023Linear} must be constructed, wherein the parameters of each layer are the weighted sum of the multimodal block and the unimodal block with learnable coefficients.

Based on the analysis above, we propose a novel framework for multimodal learngene extraction from CLIP, termed as \textbf{M}ulti\textbf{m}odal \textbf{L}earn\textbf{g}ene (\textbf{MM-LG}). As illustrated in Fig.~\ref{fig:main_method}, MM-LG comprises two distinct stages:
1) \textbf{Extraction}: 
We construct an auxiliary model structure to extract the learngene from the pre-trained CLIP with the distillation technique~\cite{hinton2015distilling}. The extracted learngene is composed of blocks for multimodal, vision and language respectively, and coefficients, which are arranged in a weighted-sum manner within the extraction. 
2) \textbf{Initialization}:
We employ the extracted learngene to numerically initialize descendant models with diverse modalities and scales by integrating the coefficients and the blocks in a weighted-sum manner to obtain the initial parameter values of each layer.
In this way, MM-LG effectively extracts the generalizable components among modalities of CLIP and initializes the descendant models with diverse scales to handle various multimodal or unimodal downstream tasks.

Our learngene extraction technique enables descendant models to achieve superior generalization capabilities across both multimodal and unimodal tasks at various model scales. 
To validate its effectiveness, we conduct comprehensive experiments across three distinct tasks: image classification, cross-modal retrieval, and image captioning. 
1) In comparison with existing learngene approaches, our method demonstrates notable improvements, achieving performance gains of 3.1\% on Oxford-IIIT PET~\cite{parkhi2012cats} for the 8-layer configuration and 4.13\% on Flickr30k~\cite{young2014image} for the 12-layer configuration compared to TLEG~\cite{Xia2023Linear}.
2) Our approach achieves comparable and sometimes superior performance to the upper bound PT-FT (Pre-training and Fine-tuning) baseline, surpassing it by 1.9\% on Oxford-IIIT PET and 3.65\% on Flickr30k in the 12-layer setting.
3) Our method requires only around 25\% of the parameter storage of the PT-FT paradigm while enabling flexible model initialization, which reduces around 2.8$\times$ pre-training costs across different model scales.
These experimental results demonstrate our method's effectiveness in extracting generalizable components from CLIP that can be successfully applied across diverse downstream tasks.

%

Our main contributions are summarized as follows:
\begin{itemize}
    \item Our work is the first to explore multimodal generalizable knowledge, which has not been taken into account in previous Learngene studies.
    \item We introduce MM-LG, a novel framework for extracting and expanding multimodal learngene from CLIP, which captures both multimodal and unimodal generalizable knowledge from CLIP.
    \item Extensive experiments show that MM-LG outperforms existing learngene methods and achieves comparable or superior results to the PT-FT paradigm across diverse tasks storing only around 25\% of the parameters without repetitive pre-training.
\end{itemize}
\section{Related Work}
\subsection{CLIP}
Some previous multimodal works have investigated methods for establishing meaningful interactions between visual and linguistic modalities~\cite{li2023blip,jia2021scaling,zhai2023sigmoid}. CLIP \cite{radford2021learning} pioneered the establishment of representational relationships between images and text through contrastive learning. Subsequently, some recent studies \cite{zhu2023learning,lai2023clipath,bao2022vlmo} have delved further into extracting cross-modal commonalities within the framework of CLIP. ALBEF~\cite{li2021align} proposed cross-modal attention to align the image and text representations with the fusion module. CLIP-CID~\cite{Yang2024CLIPCIDEC} adapted a cluster instance method to find commonalities in multimodal data.
CLIP-KD \cite{yang2024clip} explored several techniques to enhance the performance and efficiency of lightweight CLIP models through knowledge distillation. Clipping \cite{pei2023clipping} proposed a layerwise alignment method to effectively transfer the knowledge from a large pre-trained model to a smaller model.
However, despite the efforts of these methods in exploring the path to model compression and performance improvement for CLIP models, they face a significant challenge: when the scale of downstream-task models varies, these methods invariably require repetitive large-scale pre-training. 
Compared with other existing works for CLIP, our work aims to explore multimodal generalization ability, extracting the generalizable components in CLIP and utilizing them to initialize diverse models for downstream tasks.

\begin{figure*}[htbp]
    \centering
    \includegraphics[width=\textwidth]{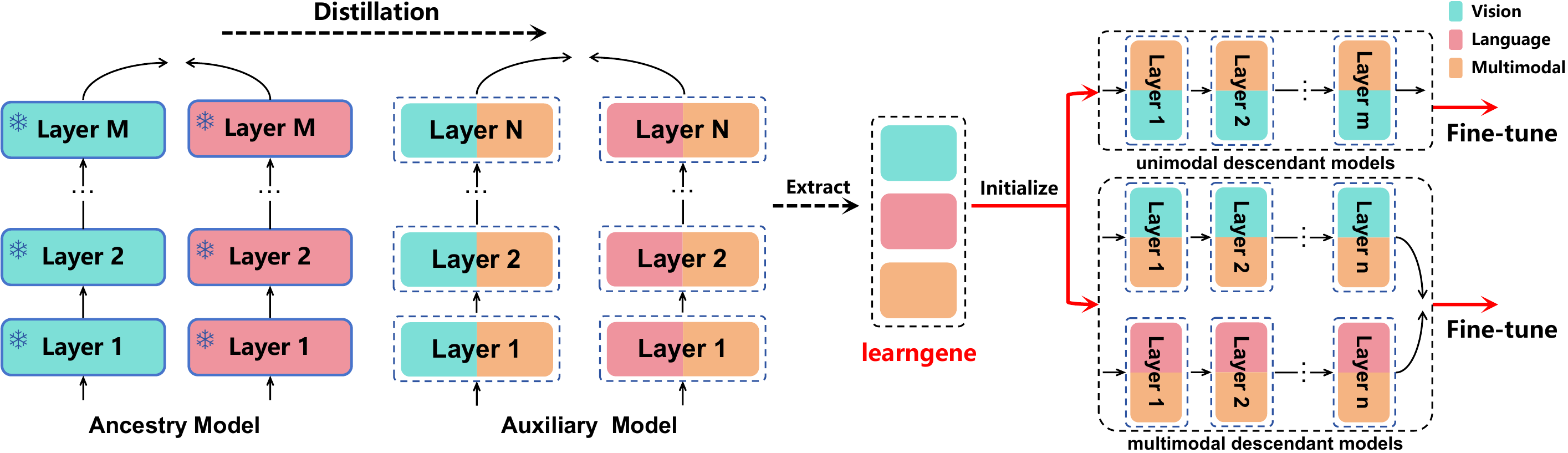}
    \caption{The framework of our proposed method comprises two stages. In the first stage, we construct an auxiliary model wherein the parameters of each layer are the weighted sum of a multimodal block and a unimodal block with learnable coefficients, and we subsequently train it through distillation against the ancestry model. After obtaining learngene composed of blocks and coefficients, in the second stage, we numerically initialize both unimodal and multimodal descendant models of varying depths, which are fine-tuned for downstream tasks.}
    \label{fig:main_method}
\end{figure*}

\subsection{Learngene}
Learngene presents an innovative and effective approach for extracting a compact yet information-rich component, termed as \textit{learngene}, from a well-trained large-scale model, known as the Ancestry Model (Ans-Net). This learngene is then utilized to initialize Descendant Models (Des-Nets) of varying sizes. Several distinct methodologies have been developed to implement this approach~\cite{wangvision,wang2023learngene,feng2024transferring,xie2024kind}.
He-LG~\cite{wang2022learngene} initially proposed extracting higher-level network layers as the learngene and integrating them with randomly initialized layers to construct Des-Nets. TLEG~\cite{Xia2023Linear} introduced a method for learngene extraction and expansion using structures with linear constraints. LearngenePool~\cite{shi2024building} employs a diverse strategy by extracting multiple small models from the Ans-Net to form learngene instances, which are then combined to create Des-Nets. SWS~\cite{xia2024exploring} constructs an auxiliary model with stage-wise weight sharing to learn the learngene, using it to initialize Des-Nets of varying sizes.
Compared with previous methods, our paradigm explores the multimodal generalizable knowledge in CLIP, extracting generalizable knowledge across modalities to initialize descendant models with varying sizes for both unimodal and multimodal tasks.


\section{Approach}

Fig.~\ref{fig:main_method} illustrates the pipeline of our proposed method, which consists of two distinct stages. In the first stage, our focus lies in extracting learngene from the ancestry model with the auxiliary model. The auxiliary model is structured with two groups of blocks and a series of coefficients, and notably, these are referred to as learngene.
For the second stage, having successfully obtained the well-extracted learngene from stage $1$, we leverage it to numerically initialize descendant models of varying sizes and diverse modalities.
Subsequently, we would delve into the details of our employed techniques.

\begin{figure*}[htbp]
    \centering
    \includegraphics[width=\textwidth]{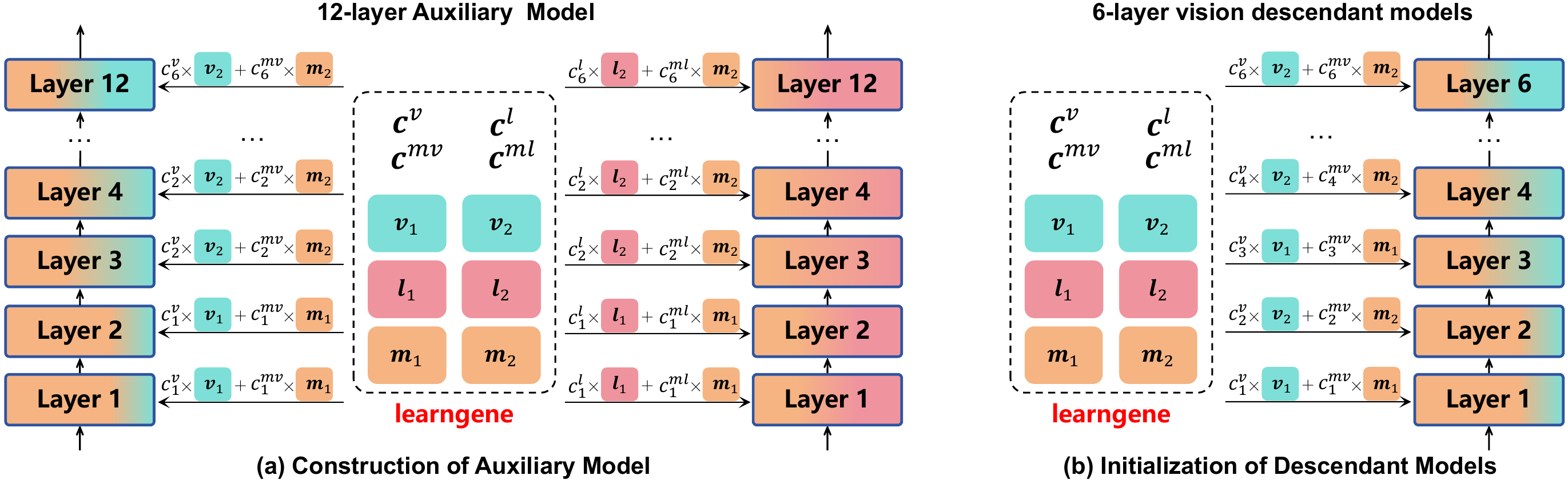}
    \caption{(a) For a 12-layer auxiliary model, during its construction, two groups of blocks $\boldsymbol{\theta}_1,\boldsymbol{\theta}_2$ are alternately and repeatedly utilized. Each group is shared twice before switching to the other. For each vector in $\boldsymbol{c}_{LG}$, every scalar weight element is shared twice before proceeding to the next one. (b) With the learngene group index and coefficient index selection in the 12-layer auxiliary model eliminating repetitions, with well-extracted learngene, the modified index selection would serve as a numerical initialization for a normal Transformer-based 6-layer vision descendant model.}
    \label{fig:detail_method}
\end{figure*}

\subsection{Construction of Auxiliary Model}
\label{sec:construct_aux}
We would delve into how to construct the auxiliary model to extract multimodal learngene in CLIP. Given that CLIP is a multimodal model with dual architecture, it follows that the components of the learngene ought to incorporate at least these two modalities. Additionally, in CLIP, there is cross-modal interaction during contrastive learning based on the outputs of these two encoders. Consequently, an additional multimodal part will be incorporated into our learngene. 
To be more specific, our leargene consists of a block part $\boldsymbol{\theta}_{LG}$ and a coefficient part $\boldsymbol{c}_{LG}$, which are shared across the auxiliary model. Here for model structure optimization and facilitate harmonious interactions, $\boldsymbol{\theta}_{LG}$ are divided into $2$ groups $\boldsymbol{\theta}_1$ and $\boldsymbol{\theta}_2$, where $\boldsymbol{\theta}_1=\{\boldsymbol{\theta}^{v}_1,  \boldsymbol{\theta}^{l}_1, \boldsymbol{\theta}^{m}_1\}$ and $\boldsymbol{\theta}_2=\{\boldsymbol{\theta}^{v}_2, \boldsymbol{\theta}^{l}_2, \boldsymbol{\theta}^{m}_2\}$. Here $\boldsymbol{\theta}^{v}, \boldsymbol{\theta}^{l}, \boldsymbol{\theta}^{m}$ refer to the blocks for vision, language and multimodal respectively and the subscript is the group index. Each $\boldsymbol{\theta}$ represents the entirety of weights and biases of linear processes for multi-head self-attention (MSA) and multi-layer perceptron (MLP) in a Transformer layer~\cite{Vaswani2017Transformer}. $\boldsymbol{c_{LG}}$ consists of $3$ parts $\boldsymbol{c}^v, \boldsymbol{c}^l$ and $\boldsymbol{c}^{m}$, where $\boldsymbol{c}^{m}=\{\boldsymbol{c}^{mv}, \boldsymbol{c}^{ml}\}$.   $\boldsymbol{c}^v, \boldsymbol{c}^l, \boldsymbol{c}^{mv},\boldsymbol{c}^{ml}$ refer to the coefficient vectors for vision, language, multimodal-vision, and multimodal-language respectively with half the length of the layer sequence.

Subsequently, we proceed to the construction part of the auxiliary model, the details of which are illustrated in Fig.~\ref{fig:detail_method}(a). During the process of constructing an auxiliary model layer, two groups of blocks $\boldsymbol{\theta}_1,\boldsymbol{\theta}_2$ are used alternately and repeatedly. That is, one group is shared twice and then switched to the other, with them taking turns in this way. For each vector in $\boldsymbol{c}_{LG}$, with its length half the number of layers, every weight scalar element is shared twice prior to the next one. Each time the group of learngene blocks and the weight scalars are confirmed, a Transformer layer is constructed. For layer $i$ in the vision encoder, the block group index $j$ and the coefficient index $k$ could be worked out with $i$ and the mentioned arrangement. Considering parameters $\boldsymbol{P}_i^v$ in Transformer layer $i$, we have Eq.~\ref{eq:Img}.
\begin{equation}
    \boldsymbol{P}_i^v=c_{k}^v \times \boldsymbol{\theta}^{v}_{j} + c^{mv}_{k} \times \boldsymbol{\theta}^{m}_{j} \label{eq:Img}
\end{equation}
For layer $i$ in the language encoder, we similarly have Eq.~\ref{eq:Txt}.
\begin{equation}
        \boldsymbol{P}_i^l=c_{k}^l \times \boldsymbol{\theta}^{l}_{j} + c^{ml}_{k} \times \boldsymbol{\theta}^{m}_{j} \label{eq:Txt}
\end{equation}\
Take layer $6$ for instance, the block group index is $2$ and the coefficient index is $3$. Thus $\boldsymbol{P}_6^v$ for vision branch is $\boldsymbol{P}_6^v=c^v_{3} \times \boldsymbol{\theta}^v_{2} + c^{mv}_{3} \times \boldsymbol{\theta}^m_{2}$ while for language branch is $\boldsymbol{P}_6^l=c^l_{3} \times \boldsymbol{\theta}^l_{2} + c^{ml}_{3} \times \boldsymbol{\theta}^m_{2}$.

Except for the parameters referred above, other components like Layer Normalization(LN)~\cite{Ba2016LayerN} part or Embedding part are not decomposed. The Layer Normalization part is shared in each Transformer layer.

\subsection{Training of Auxiliary model}
\label{sec:training_aux}
To extract the learngene in CLIP, we take advantage of knowledge distillation~\cite{hinton2015distilling} to train the auxiliary model. Combined with the learngene and components other than the learngene, a dual architecture auxiliary model can already be established. Subsequently, the output of this model can be exploited to conduct distillation on CLIP.

Before discussing the details, we abbreviate the auxiliary model as $aux$ and the ancestry model as $anc$. Assume $\{(I_k,T_k)\}_{k=1}^{|\mathcal{B}|}$ is a mini-batch of image-text pair data, with the vision and language encoders in the auxiliary model $f_v^{aux}(\cdot)$ and $f_l^{aux}(\cdot)$, $d$-dimension features $v_k^{aux}=f_v^{aux}(I_k)$, $s_k^{aux}=f_l^{aux}(T_k)$ are obtained.
Subsequently, $l2$ normalization is applied for all features. Following the implementation in ~\cite{radford2021learning}, we firstly obtain the pairwise cosine similarity logit $logit^{aux}$
\begin{equation}
   logit^{aux}=V^{aux}{S^{aux}}^{\top}/\tau^{aux}
\end{equation}
where $V^{aux},S^{aux}\in \mathbb{R}^{|\mathcal{B}|\times d} $ are feature batches for vision and language while $\tau^{aux}$ is a learnable temperature. Similarly, $logit^{anc}$ could be worked out where $\tau^{anc}$ is frozen.
Associated with the Cross-Entropy function $CE(\cdot)$ and an identity matrix $\mathbf{I}\in \mathbb{R}^{|\mathcal{B}|\times |\mathcal{B}|}$, Contrastive Language-Image Pre-Training would be performed with Eq.~\ref{eq:clip}
\begin{equation}
   \mathcal{L}_{CLIP}=\frac{1}{2}(CE(logit^{aux}, \mathbf{I})+CE({logit^{aux}}^{\top}, \mathbf{I}))
   \label{eq:clip}
\end{equation}

Except for $\mathcal{L}_{CLIP}$, the Cross-Entropy function is applied for the distillation part. With the differences calculated, the auxiliary model could effectively extract learngene from the ancestry model. The Cross-Entropy function is performed between two logits $logit^{aux}$ and $logit^{anc}$
\begin{equation}
    \mathcal{L}_{dist}^{i2t} = CE(logit^{aux}, logit^{anc})
\end{equation}
\begin{equation}
    \mathcal{L}_{dist}^{t2i} = CE({logit^{aux}}^{\top}, {logit^{anc}}^{\top})
\end{equation}
Work out the average of them to get the distillation loss
\begin{equation}
    \mathcal{L}_{dist} = \frac{1}{2}(\mathcal{L}_{dist}^{i2t} + \mathcal{L}_{dist}^{t2i})
\end{equation}
Additionally, to balance $\mathcal{L}_{CLIP}$ and $\mathcal{L}_{dist}$, a loss weight $\lambda$ is used to integrate these two losses
\begin{equation}
    \mathcal{L}_{train} = \mathcal{L}_{CLIP} + \lambda \mathcal{L}_{dist}
    \label{eq:training_loss}
\end{equation}

\subsection{Initialization of Descendant Models}

For this part, we employ the extracted learngene to make a numerical initialization for parameter values in a normal unimodal Transformer model or a normal CLIP model, as shown in Fig.~\ref{fig:detail_method}(b). We would discuss how to initialize a normal unimodal Transformer model and the process for a normal CLIP model is to combine the vision part with the language part.

\begin{table*}[htbp] 
\setlength{\tabcolsep}{1.5em}      
\centering
\caption{Performance of cross-modal retrieval on COCO and Flickr30k datasets.}
\begin{tabular}{l c c c c c}
\toprule
\multirow{2}{*}{\textbf{Method}} & \multirow{2}{*}{\textbf{Layers}} & \multicolumn{2}{c}{\textbf{COCO}~\cite{lin2014microsoft}} & \multicolumn{2}{c}{\textbf{Flickr30k}~\cite{young2014image}}\\
\cmidrule(lr){3-4} \cmidrule(lr){5-6}
& & \textbf{I2T} & \textbf{T2I} & \textbf{I2T} & \textbf{T2I} \\
\midrule
PT-FT & \multirow{5}{*}{12} & 30.56 & 30.10 & 61.53 & 59.86 \\
Scratch &  & 0.46 & 0.54 & 0.29 & 0.49 \\
He-LG~\cite{wang2022learngene} &  & 1.70 & 1.54 & 3.45 & 3.25 \\
TLEG~\cite{Xia2023Linear} &  & 29.28 & 26.56 & 61.05 & 57.89 \\
\rowcolor{gray!25}
MM-LG (Ours) &  & \textbf{33.06} & \textbf{31.48} & \textbf{65.18} & \textbf{65.18}  \\
\midrule
PT-FT & \multirow{5}{*}{8} & 28.74 & 27.50 & 58.28 & 57.20  \\
Scratch &  & 0.80 & 0.90 & 1.28 & 0.49 \\
He-LG~\cite{wang2022learngene} &  & 1.46 & 1.36 & 2.66 & 3.06  \\
TLEG~\cite{Xia2023Linear} &  & 27.22 & 24.94 & 57.30 & 55.23  \\
\rowcolor{gray!25}
MM-LG (Ours) &  & \textbf{29.08} & \textbf{27.54} & \textbf{59.37} & \textbf{57.98} \\
\midrule
PT-FT & \multirow{5}{*}{6} & \textbf{26.50} & \textbf{25.76} & \textbf{52.76} & \textbf{53.55}  \\
Scratch &  & 0.84 & 0.50 & 0.39 & 0.59  \\
He-LG~\cite{wang2022learngene} &  & 1.30 & 1.34 & 2.96 & 3.25 \\
TLEG~\cite{Xia2023Linear} &  & 24.02 & 24.40 & 47.83 & 49.11 \\
\rowcolor{gray!25}
MM-LG (Ours) &  & 25.20 & 23.54 & 51.09 & 52.17 \\
\bottomrule
\end{tabular}
\label{tab:retrieval}
\end{table*}

To initialize a unimodal Transformer descendant model, components other than the learngene could be loaded directly like Embedding or Layer Normalization. Similar to the process in auxiliary model construction, the key is to select the appropriate learngene group and coefficients to calculate the parameter values, which serves as a numerical initialization for parameter values in normal models. The selection for the learngene group and coefficients is repetitive every $2$ layers in Fig.~\ref{fig:detail_method}(a). By eliminating such repetitions, with the modified index selection in Fig.~\ref{fig:detail_method}(b), the numerical result obtained by the weighted summation of each layer can serve as a numerical initialization for parameter values in $6$-layer descendant models. When constructing a descendant model with $n$ layers, with the initialized $6$-layer descendant model, we merely require to replicate each of the first $n-6$ layers once and then connect them in sequence to accomplish the initialization. Take a $8$-layer normal Vision Transformer model~\cite{dosovitskiy2020image} for instance, the first $2$ layers of a $6$-layer descendant model are replicated like those in the auxiliary model and connected with the remaining $4$ layers to perform the $8$-layer initialization.

Except for this parameter initialization, Layer Normalization for each layer is initialized with the weight-shared one we have referred in Section~\ref{sec:construct_aux}. 
After a minor activation~\cite{muralidharan2024compact} with a small amount of data, the final model could be employed for further fine-tuning in downstream tasks. More details about the activation would be presented in Appendix~\ref{appendix:training_details}.
\section{Experiments}
In this section, we mainly explore the following questions:
\begin{enumerate}
    \item Whether MM-LG can handle multimodal downstream tasks with the extracted learngene.
    \item Whether MM-LG can handle unimodal downstream tasks with the extracted learngene.
    \item How MM-LG improves the efficiency.
\end{enumerate}

\subsection{Experimental Setup}

\textbf{Datasets.} For the learngene extraction and the pre-training, we utilize Conceptual Captions 12M (CC12M)~\cite{changpinyo2021conceptual} and Conceptual Captions 3M (CC3M)~\cite{sharma2018conceptual}.  MSCOCO~\cite{lin2014microsoft} and Flickr30k~\cite{young2014image} datasets are employed for cross-modal downstream tasks. As for visual tasks, we use CIFAR-100~\cite{krizhevsky2009learning}, Food-101~\cite{bossard2014food}, and Oxford-IIIT PET~\cite{parkhi2012cats} datasets.

\textbf{Baselines.} Scratch randomly initializes the weight and trains on the downstream dataset. The PT-FT paradigm (Pre-training and Fine-tuning) first conducts pre-training on large-scale datasets and then performs fine-tuning on downstream datasets. We extend the He-LG~\cite{wang2022learngene} method from unimodal to multimodal, and extract  learngene from a distilled ViT-S CLIP to initialize the downstream models. Similarly, we extend TLEG~\cite{Xia2023Linear} to a multimodal model, impose linear constraints on each modality to extract learngene and expand it in downstream tasks.

\textbf{Training details.} We follow the consistent training details with CLIP-related works~\cite{yang2024clip,ilharco_gabriel_2021_5143773,cherti2023reproducible}. Specifically, we extract a CLIP-ViT-S-sized auxiliary model from the pre-trained CLIP-ViT-B model, and then initialize descendant models following the requirements of downstream tasks. The experiments run over 8 Ascend 910B NPUs (64GB) hardware with RAM 1000GB.
%
%
Please refer to Appendix~\ref{appendix:training_details} for more details.

\begin{figure*}[htbp]
    \centering
    \includegraphics[width=\textwidth]{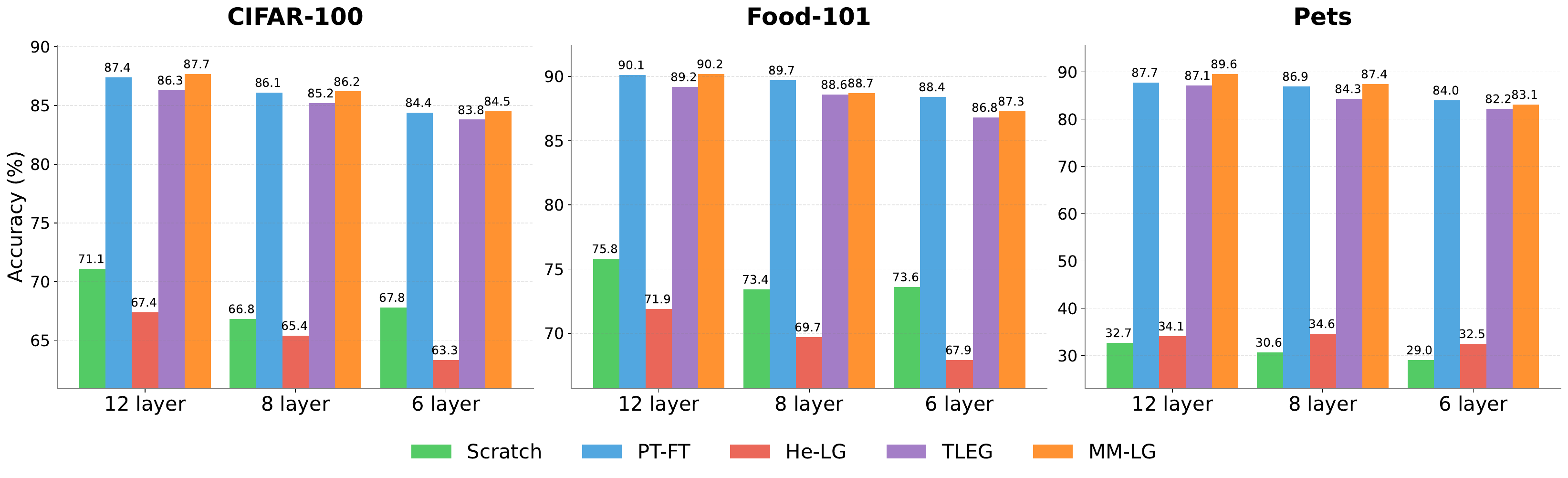}
    \caption{Performance comparison across layers and methods on CIFAR-100, Food-101, and Oxford-IIIT PET datasets.}
    \label{fig:performance}
\end{figure*}

\subsection{Cross-modal Retrieval}

The results for the cross-modal retrieval task are shown in Table~\ref{tab:retrieval}, and it is apparent that our paradigm MM-LG could handle this task with generalizable multimodal learngene.

Our framework has advantages under diverse layer settings and dataset settings compared to other learngene paradigms. Retrieval tasks are hard to handle without a sufficient initialization, thus downstream models trained from scratch manifest poor performance. He-LG also exhibits inferior performance since most layers in descendant models are directly fine-tuned without a sufficient initialization. In contrast, TLEG demonstrates normal performance but MM-LG presents substantially better performance than it. Take the image-to-text retrieval task for instance, MM-LG outperforms TLEG by \textbf{3.78\%}, \textbf{1.86\%} and \textbf{1.18\%} respectively in the 12-layer, 8-layer and 6-layer settings on COCO, while \textbf{4.13\%}, \textbf{2.07\%} and \textbf{3.26\%} in the same settings on Flickr30k. This result emphasizes the significance of constructing a multimodal block in learngene explicitly, which handles multimodal generalizable knowledge more effectively than employing linear expansion in both encoders respectively.

Compared with the upper bound, the PT-FT paradigm, our framework surprisingly exceeds it by a margin of \textbf{2.50\%}, \textbf{0.34\%}, respectively in the 12-layer, 8-layer settings on COCO, while \textbf{3.65\%}, \textbf{1.09\%} in the same settings on Flickr30k. In the 6-layer setting, our paradigm achieves comparable results merely with a slight decrease since the PT-FT paradigm costs a lot for diverse model scales. Although not surpassing the PT-FT paradigm in all layer settings, our paradigm provides a superior initialization of generalizable multimodal knowledge without repetitive pre-training.

\subsection{Image Classification}
As illustrated in Fig.~\ref{fig:performance}, though merely applying part of the learngene, our method could still handle unimodal tasks with vision and multimodal generalizable components.

Despite the traditional unimodal task of image classification, the performance demonstrated by different learngene paradigms varies substantially. For this task, the effect of negative transfer ~\cite{Wang2019CVPR} occurs to He-LG. The trained parameters of the last three layers of the ancestry model actually impair the performance of the descendant models, whose performance is inferior to that of the models trained from scratch on CIFAR-100 and Food-101. During the extraction phase, He-LG is unable to effectively extract both unimodal and multimodal generalizable knowledge, failing to handle unimodal tasks. Compared with TLEG, across all layer settings, MM-LG has an improvement of \textbf{0.7\%$\sim$ 1.4\%} on CIFAR-100 and \textbf{0.1\%$\sim$ 1.0\%} on Food-101. Even with the small-scale dataset Oxford-IIIT PET, MM-LG yet improves \textbf{2.5\%}, \textbf{3.1\%} and \textbf{0.9\%} gains over TLEG, which adequately demonstrates that MM-LG, while handling multimodal generalizable knowledge, effectively preserves unimodal generalizable knowledge. These two components handle unimodal tasks remarkably well through their cooperation in the initialization.

\begin{table}[htbp]
\setlength\tabcolsep{3pt}   
\centering
\caption{Performance of image captioning on COCO.}
\begin{tabular}{l c c c c c c}
\toprule
\textbf{Method} & \textbf{Layers} & \textbf{BLEU@4}$\uparrow$ & \textbf{CIDEr}$\uparrow$ & \textbf{ROUGE-L}$\uparrow$\\
\midrule
PT-FT & \multirow{4}{*}{12} & 38.37 & 53.90 & 44.44 \\
He-LG &  & 24.36 & 14.72 & 34.70 \\
TLEG &  & 40.19 & 56.33 & 45.39 \\
\rowcolor{gray!25}
MM-LG (Ours) &  & \textbf{41.09} & \textbf{59.34} & \textbf{45.98} \\
\midrule
PT-FT & \multirow{4}{*}{8} & 36.81 & 47.61 & 42.97 \\
He-LG &  & 23.01 & 11.25 & 33.62 \\
TLEG &  & 36.43 & 45.63 & 42.52 \\   
\rowcolor{gray!25}
MM-LG (Ours) &  & \textbf{38.99} & \textbf{53.28} & \textbf{44.74} \\
\midrule
PT-FT & \multirow{4}{*}{6} & 36.54 &	44.51 & 42.67 \\
He-LG &  & 22.86 & 13.07 & 33.54  \\
TLEG &  & 33.69 & 39.87 & 41.20 \\
\rowcolor{gray!25}
MM-LG (Ours) &  & \textbf{38.16} & \textbf{49.80} & \textbf{43.74} \\
\bottomrule
\end{tabular}
\label{tab:ic}
\end{table}

Compared with the PT-FT paradigm, MM-LG surpasses it by \textbf{0.3\%}, \textbf{0.1\%} and \textbf{1.9\%} on the listed datasets in the 12-layer configuration and achieves comparable results in the other 2 layer settings. MM-LG even outperforms it by \textbf{0.5\%} on Oxford-IIIT PET in the 8-layer setting, which exhibits our advantages in initializing models of varying sizes without seriously degrading the performance. The image classification task is a traditional downstream task and our method could achieve such performance, fully demonstrating that MM-LG has comprehensively extracted the generalizable knowledge across all modalities in CLIP.

\subsection{Image Captioning}

We conduct the image captioning task following~\cite{Mokady2021ClipCapCP}. After initializing the vision encoder, we connect it with a lightweight Transformer-based mapping network and the language model, GPT-2~\cite{radford2019language}, subsequently training the mapping network with the vision encoder and GPT-2 frozen. Since no more fine-tuning is conducted for the vision encoder, models trained from scratch would not participate in the performance comparison.

\begin{table}[htbp]
\setlength\tabcolsep{2pt}   
\centering
\caption{Performance of storage efficiency, training costs and the average image-to-text accuracy on Flickr30k.}
\begin{tabular}{l c c c}
\toprule
\textbf{Method} & \textbf{Params(M)} & \textbf{GPU-hours(H)} & \textbf{Avg Acc(\%)} \\
\midrule
PT-FT & 151.4 &  1142.4 & 57.52 \\
He-LG & \textbf{10.7} &  459.6 & 3.02  \\
TLEG & 30.3 &  457.5 & 55.39 \\
\rowcolor{gray!25}
MM-LG (Ours) & 37.4 &  \textbf{407.8} & \textbf{58.55} \\
\bottomrule
\end{tabular}
\label{tab:efficiency}
\end{table}

The results are demonstrated in Table~\ref{tab:ic}. Compared with other learngene paradigms, our method has achieved a significant advantage in this task. We apply BLEU~\cite{Papineni_Roukos_Ward_Zhu_2001}, CIDEr~\cite{Vedantam_Zitnick_Parikh_2015} and ROUGE-L~\cite{Lin_Och_2004} for evaluation metrics. Taking BLEU for example, MM-LG has an improvement of \textbf{16.73}, \textbf{15.98}, \textbf{15.30} over He-LG in the 12-layer, 8-layer and 6-layer settings, and surpasses TLEG by \textbf{0.90}, \textbf{2.56}, \textbf{4.47} in the same settings. Our method also surpasses these paradigms for the other two matrices by a large margin in diverse layer settings. Although other learngene paradigms extract generalizable knowledge to some extent in CLIP, they are less comprehensive than MM-LG in exploring multimodal generalizable knowledge. Therefore, their initialized vision encoders could hardly have further collaboration with the language model, GPT-2, to generate a satisfactory caption.

To our surprise, MM-LG outperforms the PT-FT paradigm in all layer settings in this task. Especially for the CIDEr metric, MM-LG significantly improves \textbf{5.44}, \textbf{5.67}, \textbf{5.29} gains over the PT-FT paradigm even though it has sufficient pre-training in all layer settings. Compared to the PT-FT paradigm, MM-LG acquires sufficient extraction before downstream tasks, which is more specific to multimodal and unimodal generalizable knowledge than repetitive pre-training with large-scale image-text pairs of data.

\subsection{Efficiency}

Table~\ref{tab:efficiency} presents the storage efficiency and the training costs before conducting downstream tasks and the average image-to-text performance on Flickr30k for the PT-FT paradigm and Learngene paradigms. Compared with the PT-FT paradigm, within the aspect of storage, our method merely requires 37.4M of parameter storage while the PT-FT paradigm requires 151.4M for three scales of models. We have achieved a reduction of approximately \textbf{75\%} in storage. Furthermore, for training costs our paradigm demands 407.8 GPU-hours to accomplish the learngene extraction. In contrast, the PT-FT paradigm demands 1142.4 GPU-hours, nearly \textbf{2.8$\times$} of ours because it demands a total pre-training for each model scale requirement. Besides, MM-LG's average performance outperforms the PT-FT paradigm by \textbf{1.03\%}. It significantly highlights the superiority of MM-LG in efficiency that this paradigm eliminates the need for repetitive pre-training for diverse model scales while effectively extracting the generalizable components in CLIP. 

\begin{table}[htbp]
\centering
\caption{Ablation study on cross-modal retrieval Flickr30k dataset and image classification CIFAR-100 dataset with different methods and layer configurations}
\begin{tabular}{lcccc}
\toprule
\multirow{2}{*}{\textbf{Method}} & \multirow{2}{*}{\textbf{Layers}} & \multicolumn{2}{c}{\textbf{Flickr30k}} & \multirow{2}{*}{\textbf{CIFAR-100}} \\
\cmidrule(lr){3-4}
& & \textbf{I2T} & \textbf{T2I} & \\
\midrule
\textit{w/o} MM & \multirow{3}{*}{12} & 23.47 & 22.88 & 78.11 \\
\textit{only} MM &  & 23.37 & 22.98 & 81.15 \\
\rowcolor{gray!25}
MM-LG &  & \textbf{65.18} & \textbf{65.18} & \textbf{87.70} \\
\midrule
\textit{w/o} MM & \multirow{3}{*}{8} & 18.34 & 19.23 & 78.15 \\
\textit{only} MM & & 18.44 & 19.13 & 78.81 \\
\rowcolor{gray!25}
MM-LG & & \textbf{59.37} & \textbf{57.98} & \textbf{86.24} \\
\midrule
\textit{w/o} MM & \multirow{3}{*}{6} & 13.12 & 13.12 & 75.11\\
\textit{only} MM & & 12.92 & 13.81 & 76.79 \\
\rowcolor{gray!25}
MM-LG & & \textbf{51.09} & \textbf{52.17} & \textbf{84.50} \\
\bottomrule
\end{tabular}
\label{tab:ablation}
\end{table}

In comparison with the previous Learngene paradigms, the training costs are analogous to theirs since Learngene paradigms need merely a single extraction procedure before downstream tasks. The parameter storage for MM-LG is marginally larger compared to the other two paradigms due to the demand for preserving both multimodal and unimodal generalizable knowledge, leading to superior downstream performance.

\subsection{Ablation Study}

In the introduction section, it has been analyzed that there are a multimodal block, unimodal blocks and corresponding coefficients extracted in the first stage, representing the generalizable components for multimodal and unimodal modalities respectively. We conduct this ablation to verify whether the superior performance of MM-LG is due to one generalizable component alone or their collaboration.

The results are shown in Table~\ref{tab:ablation}, where "\textit{w/o} MM" denotes initializing descendant models without the extracted multimodal part and "\textit{only} MM" denotes initializing descendant models without the extracted unimodal part. For downstream multimodal tasks, we take image-to-text retrieval on Flickr30k for instance, MM-LG surpasses them by a large margin of about \textbf{42\%}, \textbf{41\%}, \textbf{38\%} in the 12-layer, 8-layer, 6-layer settings accordingly, while for unimodal tasks, we outperform them by about \textbf{8\%}, \textbf{8\%}, \textbf{9\%} on CIFAR-100 in the same layer settings. It is manifest that these two generalizable components are required to provide a superior initialization in concert, which is coherent with the cross-modal generalization ability in CLIP.
\section{Conclusion}
In this paper, inspired by employing a multimodal block to preserve multimodal generalizable knowledge in CLIP, we propose a novel and effective learngene extraction method, termed as ~\LG. \LG~is capable of extracting the generalizable knowledge in both multimodal and unimodal modalities from the learngene extraction stage, initializing descendant models for diverse downstream scenarios with different depth and modality requirements. The experimental results verify that \LG~ surpasses other learngene paradigms and achieves comparable or superior performance to the pre-training and fine-tuning paradigm without repetitive pre-training.
\section*{Ethical Statement}

There are no ethical issues.



\bibliographystyle{named}

\clearpage
\appendix
\onecolumn
\section{Implementation Details}
\label{appendix:training_details}
MM-LG is mainly divided into two stages: 1) Pre-training for extraction; 2) Initialization and fine-tuning.

\subsection{Pre-training for Extraction}
As detailed in Section~\ref{sec:construct_aux}, we develop a lightweight CLIP auxiliary model based on the ViT-S architecture. This auxiliary model features a hidden dimension of 384 and employs 6 attention heads in its multi-head self-attention mechanism. To enhance the model's capabilities, we leverage knowledge distillation techniques to transfer learngene from a larger pre-trained CLIP model based on the ViT-B architecture\footnote{\url{https://huggingface.co/timm/vit\_base\_patch16\_clip\_224.openai}}. The distillation loss for the auxiliary model is provided in Eq.~\ref{eq:training_loss} with the loss weight $\lambda$ set to $1$. Through our proposed extraction methodology, we successfully isolate three key components: (1) the multimodal learngene that captures generalizable knowledge across modalities, (2) modality-specific learngene that preserves unimodal generalizable knowledge of vision and language respectively, and (3) the corresponding importance coefficients that determine the contribution of each learngene block.

The training process is conducted on a combined dataset consisting of CC12M~\cite{changpinyo2021conceptual} and CC3M~\cite{sharma2018conceptual}, spanning 32 epochs to ensure convergence.
Our final 12-layer auxiliary model demonstrates strong performance across multiple evaluation metrics. In zero-shot image classification on the ImageNet benchmark~\cite{deng2009imagenet}, the trained auxiliary model achieves a top-1 accuracy of 41.71\%. On the CC3M validation set, the model exhibits balanced cross-modal retrieval capabilities, attaining a text-to-image retrieval (T2I) recall@1 of 37.18\% and an image-to-text retrieval (I2T) recall@1 of 37.33\%.

\subsection{Initialization and Fine-tuning}
\subsubsection{Parameter Activation}
The activation of parameters is a crucial step when applying learngene for initializing downstream models of varying scales, a phenomenon also observed in ~\cite{muralidharan2024compact}. In our implementation, we discover that a minimal activation phase using just 10\% of the original pre-training dataset for a single epoch is sufficient to effectively initialize the parameters.

To systematically investigate the relationship between activation data volume and downstream task performance, we conduct a comprehensive set of experiments. The results, presented in Table~\ref{tab:activation}, reveal a remarkable finding: our MM-LG approach requires only a fraction of the activation data to achieve performance levels comparable to traditionally pre-trained and fine-tuned (PT-FT) models. This efficiency in data utilization demonstrates the robustness and practicality of our method, particularly in scenarios where computational resources or training data are limited.

\begin{table*}[htbp]
\centering
\setlength{\tabcolsep}{1.5em}      
\caption{Performance on Flickr30k and CIFAR-100 by utilizing 10\% to 50\% of the data to activate}
\begin{tabular}{lcccc}
\toprule
\multirow{2}{*}{\textbf{Methods}} & \multirow{2}{*}{\textbf{Layers}} & \multicolumn{2}{c}{\textbf{Flickr30k}~\cite{young2014image}} & \multirow{2}{*}{\textbf{CIFAR-100}~\cite{krizhevsky2009learning}} \\
\cmidrule(lr){3-4}
& & \textbf{I2T} & \textbf{T2I} & \\
\midrule
PT-FT & \multirow{4}{*}{8} & 58.28 & 57.20 & 86.10 \\
MM-LG (10\%) &  & \textbf{59.37} & \textbf{57.98} & 86.24 \\
MM-LG (30\%) &  & 58.48 & 56.61 & 86.12 \\
MM-LG (50\%) &  & 57.40 & 56.41 & \textbf{86.37} \\
\midrule
PT-FT & \multirow{4}{*}{6} & 52.76 & \textbf{53.55} & 84.37 \\
MM-LG (10\%) &  & 51.09 & 52.17 & \textbf{84.50} \\
MM-LG (30\%) &  & \textbf{53.16} & 52.27 & 84.29 \\
MM-LG (50\%) &  & 52.37 & 51.17 & 84.39 \\
\bottomrule
\end{tabular}
\label{tab:activation}
\end{table*}

\subsubsection{Fine-tuning}
MM-LG demonstrates versatility across both multimodal and unimodal tasks, with architectures available in three scales: 12-layer, 8-layer, and 6-layer configurations. We conduct comprehensive evaluations across multiple downstream tasks:

\begin{itemize}
    \item First, for cross-modal retrieval, we evaluate the model's performance through extensive training on two benchmark datasets: MSCOCO and Flickr30k, with training conducted over 50 epochs.
    \item For the task of Image Classification, we perform fine-tuning experiments on three diverse datasets: CIFAR-100, Food-101, and Oxford-IIIT PET. Each dataset undergoes thorough training for 500 epochs to ensure complete convergence.
    \item To assess Image Captioning capabilities, we conduct extensive testing on the MSCOCO dataset, which serves as a standard benchmark in the field.
\end{itemize}

The experimental results demonstrate that our approach significantly reduces computational resource requirements while maintaining performance levels comparable to the PT-FT paradigm baseline. This achievement represents a meaningful advancement in efficient model training and deployment.

\subsection{Hyperparameters}

\begin{table}[htbp]
\centering
\setlength{\tabcolsep}{0.8em}  
\caption{Hyperparameters for MM-LG Extraction and Fine-tuning}
\begin{tabular}{lcccc}
\toprule
\multirow{2}{*}{\textbf{Hyperparameters}} & \multicolumn{4}{c}{\textbf{Training Settings}} \\
\cmidrule{2-5}
& \multicolumn{1}{c|}{\textbf{Extraction}} & \textbf{Retrieval} & \textbf{Classification} & \textbf{Captioning} \\
\midrule
Warmup steps & 10,000 & 10,000 & 5,000 & 5,000 \\
Batch size & 512 & 256 & 256 & 256 \\
Learning rate & 1e-3 & 5e-5 & 1e-4 & 2e-5 \\
Weight decay & 0.1 & 0.01 & 0.01 & --- \\
Epochs & 32 & 50 & 500 & 10 \\
\bottomrule
\end{tabular}
\end{table}




\end{document}